%% file: main.tex
\documentclass[final,5p,times,twocolumn]{elsarticle} % Final

\biboptions{sort&compress}
\graphicspath{{./figures/}}

\newcommand*{\bibliopath}{./biblio}

\usepackage{natbib}

\usepackage{mathtools}
\usepackage{amsmath}
\usepackage{amssymb}
\usepackage{bm}

\usepackage[dvipsnames]{xcolor}  % More defined colors
\usepackage{pdflscape} % for rotating to landscape page (for large tables)
\usepackage{enumitem} % to change format of enumerates
\usepackage{ragged2e}
\usepackage[hyphens]{url} % add hyphens to long url in bibliography
\usepackage[colorlinks=True, breaklinks=true]{hyperref} 
\hypersetup{allcolors=blue, pdfauthor=author}
\usepackage[nameinlink]{cleveref}

\usepackage[linesnumbered,ruled,noend]{algorithm2e} % for pseudo-code
\SetKwComment{Comment}{$\triangleright$\ }{}

\usepackage{caption}
\usepackage{subcaption}
\usepackage{float} % precise figure placement
\usepackage[export]{adjustbox} % shift a figure left or right
\usepackage{stfloats} % To enable figures at the bottom of page

\usepackage{forest}
\usepackage{tikz}
\usetikzlibrary{shapes, arrows, calc, positioning, patterns}
\usetikzlibrary{arrows.meta}
\usetikzlibrary{backgrounds}
\usepackage{booktabs} % For tables
\usepackage{multirow} %for multirow tables

\usepackage{appendix}

\journal{ArXiv}
\begin{document}
\begin{frontmatter}


\title{Disentangling Variational Autoencoders}

\author[1]{Rafael Pastrana\corref{cor1}}
\ead{arpastrana@princeton.edu}

\cortext[cor1]{Corresponding author}

\address[1]{School of Architecture, Princeton University, United States of America}
\vspace{-3cm}

%%%%%%%%%%%%%%%%%%%%%%%%%%%%%%%%%%%%%%
%% ABSTRACT
%%%%%%%%%%%%%%%%%%%%%%%%%%%%%%%%%%%%%%

\input{abstract.tex}


\end{frontmatter}
\raggedbottom
\hyphenpenalty=1000

%%%%%%%%%%%%%%%%%%%%%%%%%%%%%%%%%%%%%%
%% ARTICLE BODY
%%%%%%%%%%%%%%%%%%%%%%%%%%%%%%%%%%%%%%

\input{sections/01_intro}
\input{sections/02_background}
\input{sections/03_method}
\input{sections/04_results}
\input{sections/05_conclusion}

%%%%%%%%%%%%%%%%%%%%%%%%%%%%%%%%%%%%%%
%% BIBLIOGRAPHY
%% The .bib file is found in the folder, export from citation manager to BIBTEX

\bibliographystyle{biblio/elsarticle-num}
\bibliography{biblio/automatic_entries, \bibliopath/manual_entries}

%%%%%%%%%%%%%%%%%%%%%%%%%%%%%%%%%%%%%%
%% ARTICLE ENDING
%%%%%%%%%%%%%%%%%%%%%%%%%%%%%%%%%%%%%%

\end{document}

%% file: abstract.tex
%%%%%%%%%%%%%%%%%%%%%%%%%%%%%%%%%%%%%%
%% ABSTRACT
%%%%%%%%%%%%%%%%%%%%%%%%%%%%%%%%%%%%%%
\begin{abstract}
  % Why do we care
  A variational autoencoder (VAE) is a probabilistic machine learning framework for posterior inference that projects an input set of high-dimensional data to a lower-dimensional, latent space.
  The latent space learned with a VAE offers exciting opportunities to develop new data-driven design processes in creative disciplines, in particular, to automate the generation of multiple novel designs that are aesthetically reminiscent of the input data but that were unseen during training.
  % Problem
  However, the learned latent space is typically disorganized and entangled: traversing the latent space along a single dimension does not result in changes to single visual attributes of the data.
  The lack of latent structure impedes designers from deliberately controlling the visual attributes of new designs generated from the latent space.
  % such as the line-weight, the scale or the tilt in a digit.
  % Statement
  % To advance the adoption of machine learning-powered tools in design, it is important to develop methodologies to automatically impose a disentangled structure to the latent space.
  % that automatically align latent dimensions with interpretable dimensions or axis for  interpretability and aesthetic control in creative generative design tasks.
  % aligned with human-interpretable, visual generative factors.
  % Solution: method
  % to impose a disentangled structure to the latent space.
  % to disentangle the latent variable of a generative model such that the latent dimensions correspond to human interpretable visual design factors.
  % Ultimately, we aim to contribute to the development of methodologies that automatically align to align the latent dimensions with generative dimensions or axis for human interpretability and aesthetic control in creative generative design tasks.
  % with the goal of automatically aligning the latent dimensions with interpretable visual properties of the digits.
  This paper presents an experimental study that investigates latent space disentanglement.
  We implement three different VAE models from the literature and train them on a publicly available dataset of 60,000 images of hand-written digits.
  We perform a sensitivity analysis to find a small number of latent dimensions necessary to maximize a lower bound to the log marginal likelihood of the data.
  Furthermore, we investigate the trade-offs between the quality of the reconstruction of the decoded images and the level of disentanglement of the latent space.
  % Results
  We are able to automatically align three latent dimensions with three interpretable visual properties of the digits: line weight, tilt and width.
  Our experiments suggest that i) increasing the contribution of the Kullback-Leibler divergence between the prior over the latents and the variational distribution to the evidence lower bound, and ii) conditioning input image class enhances the learning of a disentangled latent space with a VAE.
\end{abstract}

%% file: sections/01_intro.tex
%%%%%%%%%%%%%%%%%%%%%%%%%%%%%%%%%%%%%%
%% Introduction
%%%%%%%%%%%%%%%%%%%%%%%%%%%%%%%%%%%%%%

\section{Introduction}
\label{intro}

A \textit{variational autoencoder} (VAE) is a posterior inference framework to fit latent variable models.
A VAE-fitted model has two attributes relevant to creative disciplines like type and architectural design.
% dimensionality reduction and synthetic data generation.
First, the model can project the high-dimensional input data points $\mathbf{x}$ to a lower-dimensional latent space parametrized by a latent variable $\mathbf{z}$.
Second, the learned latent space is typically a smooth and continuous manifold that can be utilized for synthetic data generation.
The latent space attains these properties because it is constrained to fit a variational probability distribution $q$ at training time.
As a result, it is possible to take samples from the VAE-fitted latent space and then decode them to generate new data points $\hat{\mathbf{x}}$ that resemble the attributes of $\mathbf{x}$ but were never part of the input data.

Powered by neural networks, VAEs have gained state of the art prominence in creative design tasks that enable the generation of synthetic but photorealistic images of people who do not exist \cite{vahdat_nvaedeep_2020}, the development of new music creation tools \cite{roberts_hierarchicalvariational_2017} and the design of novel dancing choreographies \cite{pettee_imitationgenerative_2019}.
However, the latent space spanned by $\mathbf{z}$ and learned with a VAE is typically entangled and thus lacks two important factors to operationalize it in design tasks: \textit{independence} and \textit{interpretability} \cite{higgins_definitiondisentangled_2018}.
The goal of attaining a disentangled latent space, on the other hand, is to find a low-dimensional representations of the data $\mathbf{x}$ where single latent dimensions $z_{i}$ are sensitive to changes in single generative factors while being relatively invariant to changes in others \cite{bengio_representationlearning_2013}.
To advance the use of machine learning-powered probabilistic tools in design, the desiderata is to develop and explore methodologies that produce a disentangled latent space that captures visual basic concepts imbued in the input data, such as scale, tilt, orientation or color.
This way, one can have finer and deliberate control when generating novel data from that latent space.

In this paper, we investigate three different methodologies to disentangle the latent space learned by a VAE.
First, we implement a standard VAE and replicate a portion of the numerical results in the original VAE paper on the autoencoding variational Bayes by Kingma and Welling \cite{kingma_autoencodingvariational_2014}.
Next, we explore the $\beta$-VAE \cite{higgins_betavaelearning_2016} framework, an extension to the original VAE formulations that introduces a hyperparameter $\beta$ to hypothetically enforce a semantically richer yet efficient representation of the data in the latent space.
Lastly, we investigate to what extent conditioning the dataset $\mathbf{x}$ on class labels $\mathbf{u}$ facilitates disentanglement with a $\beta$-VAE \cite{locatello_challengingcommon_2019}.
We perform multiple experiments to analyze the impact of $\beta$ on the ELBO and qualitatively assess the disentanglement of the latents.

This paper is accompanied by a blog post with interactive graphics and plots which the reader can access at \url{https://bit.ly/3KyZ0Of}.
For reproducibility purposes, we also make available a Pytorch \cite{paszke_pytorchimperative_2019} implementation of our work at \url{https://github.com/arpastrana/neu_vae}.

%% file: sections/02_background.tex
%%%%%%%%%%%%%%%%%%%%%%%%%%%%%%%%%%%%%%
%% Theoretical background
%%%%%%%%%%%%%%%%%%%%%%%%%%%%%%%%%%%%%%

\section{Theoretical background}
\label{background}

\begin{figure*}[!t]
  \centering
  \begin{subfigure}[b]{0.32\textwidth}
    \centering
    \includegraphics[width=\textwidth]{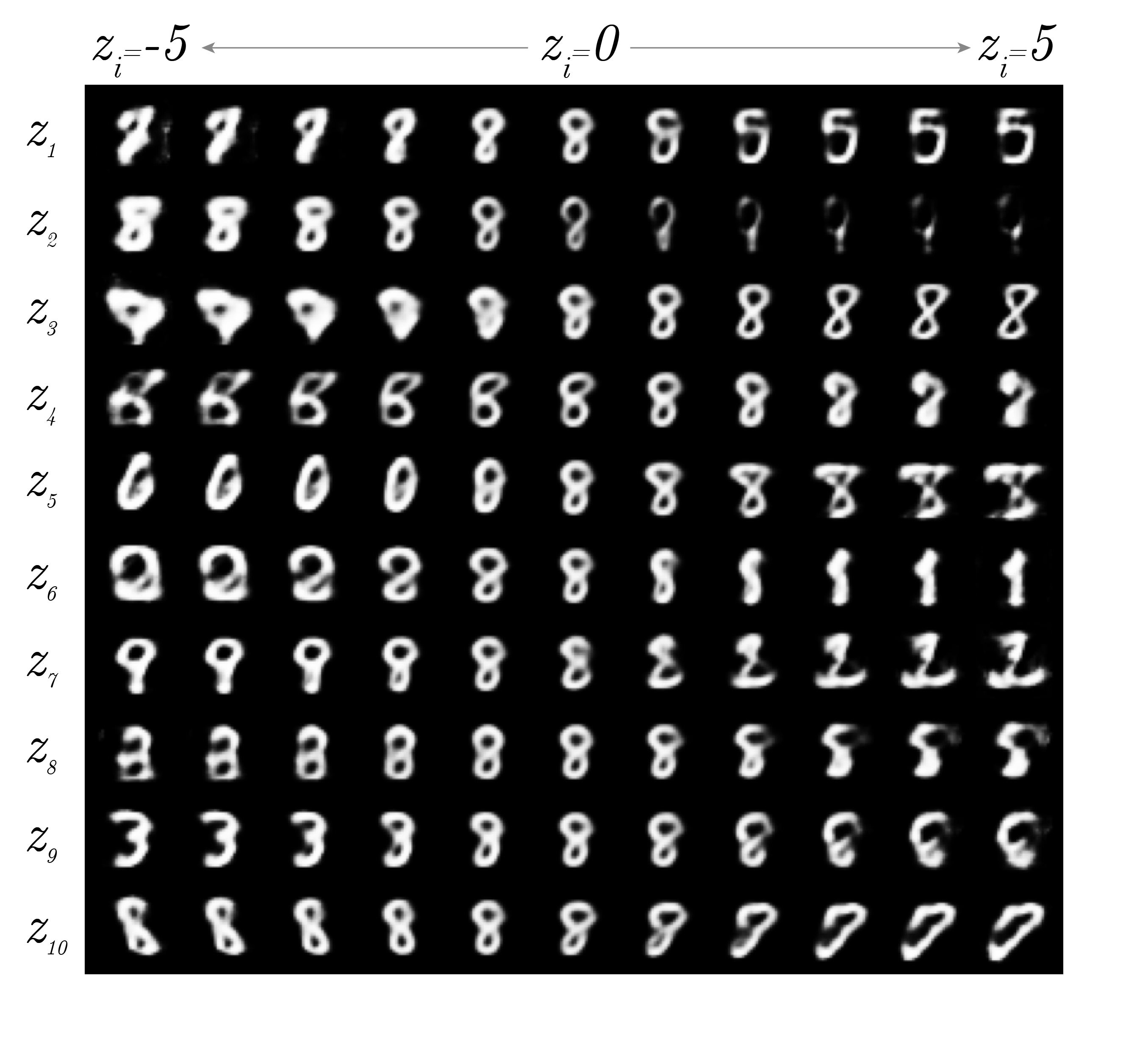}
    \caption{Standard VAE}
    \label{fig:vae:grid:8}
  \end{subfigure}
  \begin{subfigure}[b]{0.32\textwidth}
    \centering
    \includegraphics[width=\textwidth]{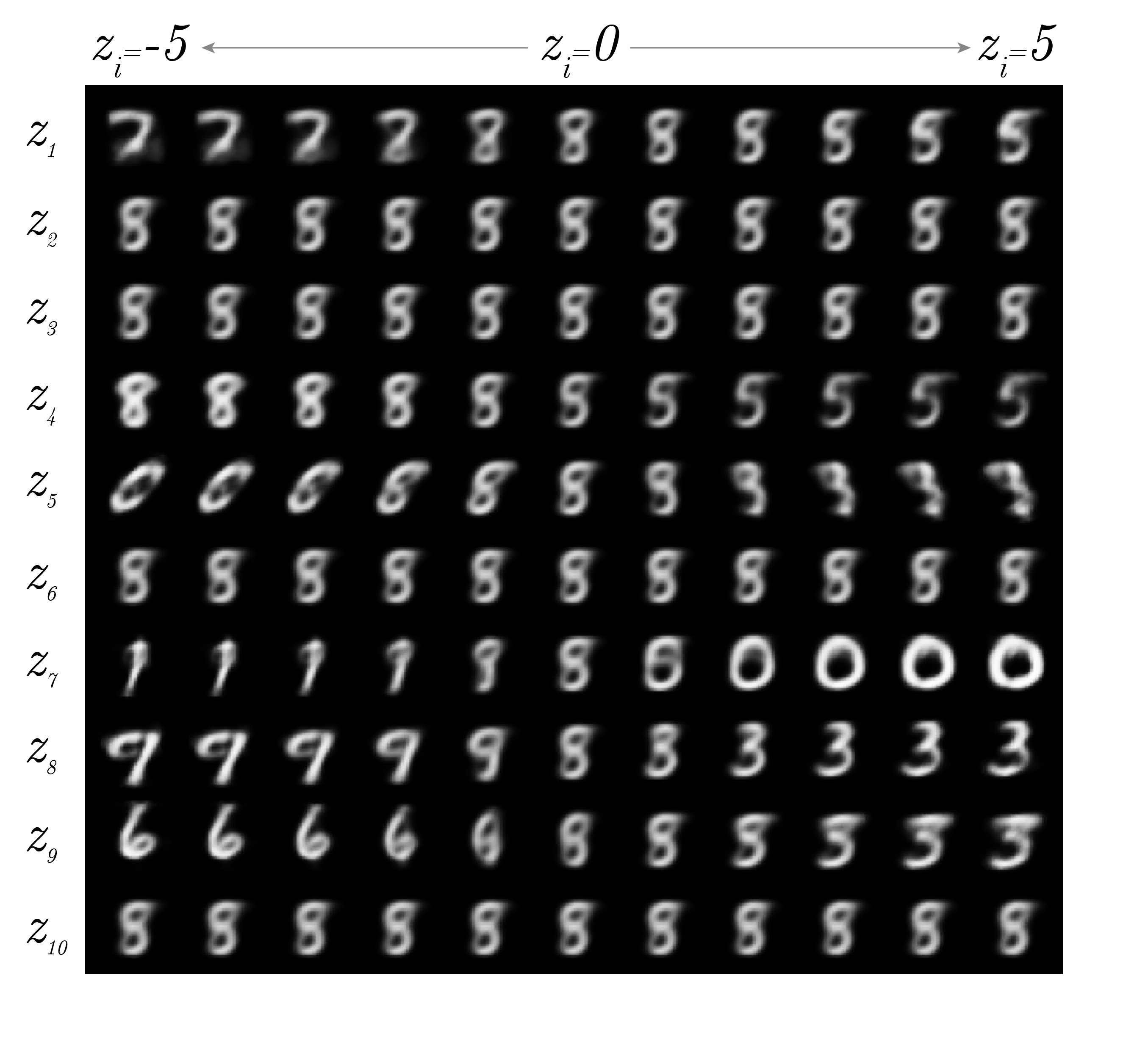}
    \caption{$\beta$-VAE, $\beta=10$}
    \label{fig:betavae:grid:8}
  \end{subfigure}
  \begin{subfigure}[b]{0.32\textwidth}
    \centering
    \includegraphics[width=\textwidth]{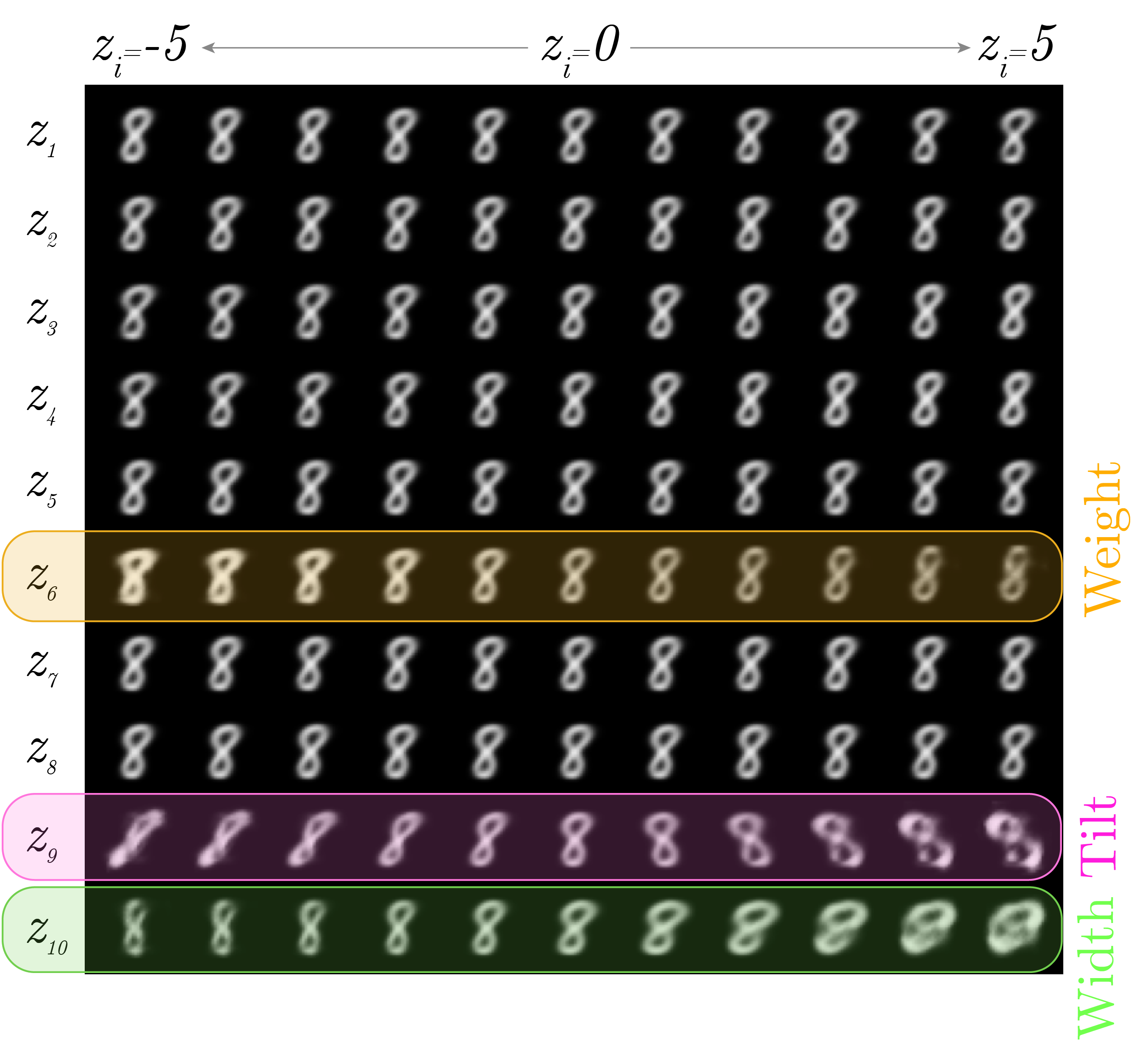}
    \caption{Conditional $\beta$-VAE, $\beta=10$}
    \label{fig:cond:betavae:grid:8}
  \end{subfigure}
\caption{\textbf{(Dis)entangled latent spaces}. 
Latent space disentanglement comparison for three different VAEs that project the input data to a 10-dimensional latent space, $J=10$.
We feed in the same image to each of the three VAEs and traverse the latent space as we describe in Section \ref{sec:disentanglement_evaluation}.
The standard VAE produces the sharpest digit reconstruction.
Only the latent space learned by conditional $\beta$-VAE shows evidence of a disentangled latent space.
Walking along latent dimensions $z_6$, $z_9$ and $z_{10}$ evidences that these dimensions align with three different visual properties: line weight, tilt and width.
}
\end{figure*}

The observed data $\mathbf{x} \in \mathbb{R}^{N}$ is a random variable that follows a probability distribution $p(\mathbf{x})$.
If we parametrize the density of the observed data with an unobserved variable $\mathbf{z} \in \mathbb{R}^{J}$, the log joint distribution of $\mathbf{x}$ and $\mathbf{z}$ is expressed as:
\begin{equation}
\log p(\mathbf{x}, \mathbf{z}) = \log p(\mathbf{x} | \mathbf{z})p(\mathbf{z})
\end{equation}

Marginalizing the joint distribution over $\mathbf{z}$ to recover the log probability of the data $\log p(\mathbf{x})$ is generally intractable.
Therefore, we maximize a lower bound to the log marginal likelihood of the data, the Evidence Lower Bound (ELBO), and fit instead the parameters of an variational distribution $q_{\phi}(\mathbf{z})$.
\begin{equation}
ELBO(\theta, \phi) = \mathbb{E}_{q_{\phi}(\mathbf{z} | \mathbf{x})} [\log p_{\theta}(\mathbf{x} | \mathbf{z})] - D_{KL}(q_{\phi}(\mathbf{z} | \mathbf{x}) || p(\mathbf{z}))    
\end{equation}

\noindent where $\phi$ and $\theta$ are the parameters of a VAE's encoder and decoder respectively.
These parameters can be computed by non-linear function approximators such as neural networks.

We can use MCMC to compute the expectation of the conditional likelihood of the data under the variational distribution, $\mathbb{E}_{q_{\phi}(\mathbf{z} | \mathbf{x})} [\log p_{\theta}(\mathbf{x} | \mathbf{z})]$.
In practice, a single sample per data point per batch suffices to approximate the expectation, which leads to the following updated expression for the ELBO:
\begin{equation}
    ELBO (\theta, \phi) \approx \log p_{\theta}(\mathbf{x} | \mathbf{z})- D_{KL}(q_{\phi}(\mathbf{z} | \mathbf{x}) || p(\mathbf{z}))
\end{equation}

The first term of this approximation can be understood as the reconstruction loss of e.g. Bernoulli- or Gaussian-decoded data points, while the second corresponds to a regularization term which penalizes the Kullback-Leibler (KL) divergence between the variational posterior $q_{\phi}(\mathbf{z} | \mathbf{x})$ and a prior $p(\mathbf{z})$ over the latent $\mathbf{z}$.
To make the optimization of the ELBO tractable and differentiable, we assume that both the posterior $q_{\phi}(\mathbf{z} | \mathbf{x})$ and the prior $p(\mathbf{z})$ are Gaussian distributions $\mathcal{N}(\mu_{j},\:\sigma_{j}^2)$ with mean $\mu_{j}$ and variance $\sigma_j^2$ per latent dimension $j$.
The choice of a Gaussian allows the calculation of the KL divergence between the variational posterior and the prior over the latents in closed form:

\begin{equation}
   \frac{1}{2} \sum_{j=1}^{J} (1 + \log((\sigma_j)^2) - (\mu_j)^2-(\sigma_j)^2)
\end{equation}

Furthermore, the Gaussian assumption enables gradient backpropagation using the \textit{reparametrization trick}, where we can sample from the variational distribution $\mathbf{z} \sim q_{\phi}(\mathbf{z} | \mathbf{x}) = \mathcal{N}(\mu,\:\sigma^2)$ via an affine transformation of an auxiliary noise variable $\epsilon$ sampled from a standard normal, $\epsilon \sim \mathcal{N}(0,\:1)$:

\begin{equation}
    \mathbf{z_{j}} = \mu_{j} + \epsilon\:\sigma_{j}
\end{equation}

\subsection{$\beta$-VAE}

To focus on learning statistically independent latent factors, the authors of \cite{higgins_betavaelearning_2016} include an additional hyperparameter $\beta$ in the loss function proposed in the original formulation of the VAE.
This hyperparameter scales the weight of the KL Divergence term in the ELBO:

\begin{equation}
    ELBO (\theta, \phi) \approx \log p_{\theta}(\mathbf{x} | \mathbf{z})- \beta D_{KL}(q_{\phi}(\mathbf{z} | \mathbf{x}) || p(\mathbf{z}))
\end{equation}

A $\beta$-VAE with $\beta=1$ corresponds to the original VAE formulation.
Employing values of $\beta > 1$ hypothetically put pressure on the VAE bottleneck to match the prior $p(\mathbf{z})$ and thus promote the learning of a more efficient latent data representation.
As reported in \cite{higgins_betavaelearning_2016}, disentanglement comes at the cost of a diminished reconstruction quality of $\mathbf{x}$.
Moreover, too small or too large $\beta$ values may not necessarily lead to disentangled latents.
Therefore, $\beta$ needs to be calibrated using qualitatively (e.g. using visual heuristics) or quantitatively approaches (e.g. developing a quantitative disentanglement metric after appending a simple linear classifier to a trained VAE).

\subsection{Conditional $\beta$-VAE}

Recent work in representation learning suggests that the latent space of fitted latent variable models cannot be disentangled without supervision \cite{locatello_challengingcommon_2019,khemakhem_variationalautoencoders_2020}.
One way to fit disentangled latent spaces is thus to build a conditionally factorized prior over the latent variables $p(\mathbf{z}|\mathbf{u})$, which is possible by concurrently observing an auxiliary variable $\mathbf{u}$ that corresponds to the time index in a time series, previous data points, or class labels \cite{khemakhem_variationalautoencoders_2020}.
This means that the model takes as input a dataset with observation pairs $\mathcal{D} = \{(\mathbf{x}_i, \mathbf{u}_i), ..., (\mathbf{x}_n, \mathbf{u}_n)\}$ instead of $\mathcal{D} = \{\mathbf{x}_i, ..., \mathbf{x}_n\}$.
The total log density of the data $\mathbf{x}$ conditioned on labels $\mathbf{u}$ is found after marginalizing the latent variables $\mathbf{z}$:

\begin{equation}
    \log p(\mathbf{x}|\mathbf{u}) = \int \log p(\mathbf{x} | \mathbf{z})p(\mathbf{z}| \mathbf{u})d\mathbf{z}
\end{equation}

%% file: sections/03_method.tex
%%%%%%%%%%%%%%%%%%%%%%%%%%%%%%%%%%%%%%
%% Methodology
%%%%%%%%%%%%%%%%%%%%%%%%%%%%%%%%%%%%%%

\section{Method}
\label{method}

\subsection{Dataset}

We train our model with the MNIST dataset \cite{lecun_mnisthandwritten_2010}.
This dataset consists of 60,000 images of ten different hand-written digits in the range $[0-9]$.
The resolution per image is of $28\times28$ pixels.
We use a randomized data loader to manage the train and validation sets that are utilized to fit the parameters of the VAE.

\subsection{VAE Architecture}

\begin{figure}[!b]
    \centering
    \begin{subfigure}[b]{0.95\columnwidth}
      \centering
      \includegraphics[width=\textwidth]{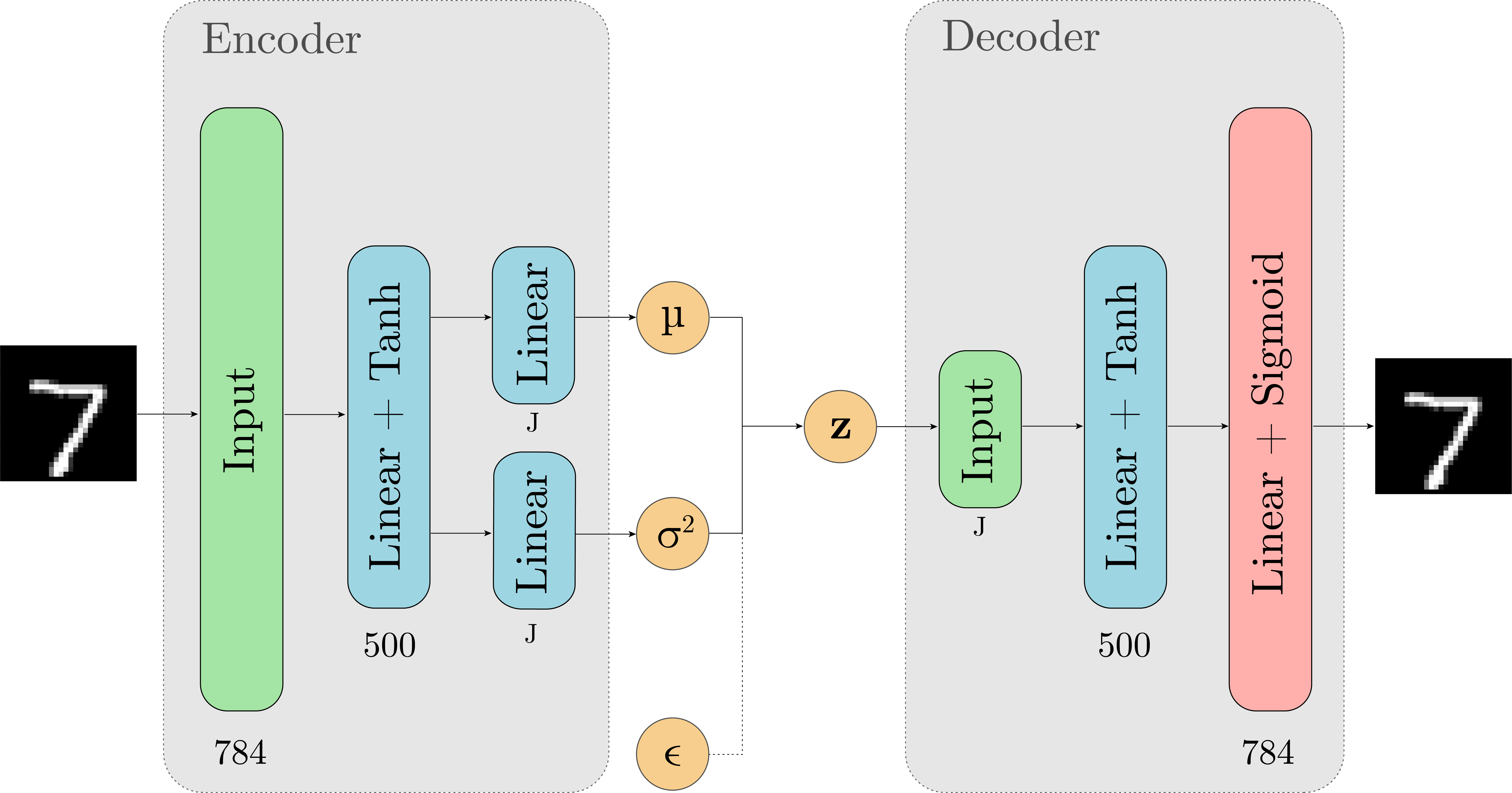}
    \end{subfigure}
    \caption{\textbf{VAE architecture}. The architecture of the VAE in all our experiments follows the structure reported in \cite{kingma_autoencodingvariational_2014}, except for when we work with a conditional $\beta$-VAE.
    The VAE consists of a three-layer encoder followed by a three-layer decoder. 
    The encoder outputs the mean $\mu$ and the variance $\sigma^2$ of the variational posterior.
    We then sample $\epsilon$ from a standard normal $\mathcal{N}(0, I)$ and use the reparametrization trick to get the latent vector $\mathbf{z}$ that is fed to the decoder.}
    \label{fig:vae:arch}
\end{figure}

For consistency, we use the same neural architecture as in Kingma and Welling \cite{kingma_autoencodingvariational_2014} in all our experiments unless otherwise noted.
We use multilayer perceptrons (MLPs) in the encoder and decoder of the VAE.
Every perceptron unit in a MLP is activated using the hyperbolic tangent function, except for the output layer in the decoder where we apply a sigmoid non-linearity.

\subsubsection{Gaussian encoder}

The encoder consists of an input linear layer with 784 units (corresponding to a flattened $28\times28$ image) followed by a fully-connected, 500-units linear layer.
To calculate the mean $\mu$ and the variance $\sigma$ of the variational distribution $q$, the output layer of the decoder is another linear layer whose number of hidden units is twice the dimensionality of $\mathbf{z}$, that is $2 \times J$.

\subsubsection{Bernoulli decoder}

After applying the reparametrization trick, we feed samples from $\mathbf{z}$ into the decoder's input linear layer which has the same number of units as in the size of the latent space, $J$.
We complete the decoder with an intermediate 500-unit linear layer followed by a 784-unit output linear layer.

\subsection{Training}

We use Adagrad \cite{duchi_adaptivesubgradient_} with a fixed learning rate of $0.01$ in all our the experiments.
We use an image batch size to 100 to compute gradients.
We cap training at 200 epochs.
In our work, the optimization objective well saturates within this budget of epochs.
The optimization objective is to minimize the negative value of the ELBO.
The reconstruction error is computed using the binary cross-entropy, and the KL divergence is calculated analytically since we assume Gaussians for the variational posterior and the prior.

\subsection{Disentanglement evaluation}
\label{sec:disentanglement_evaluation}

We traverse the $J$ distinct dimensions of the fitted $\mathbf{z}$, as in the work of Higgins, et al. \cite{higgins_betavaelearning_2016}, to visually examine the disentanglement quality of the learned latent representations.

First, we input a digit image into the VAE encoder to obtain the mean $\mu$ and the variance $\sigma$ of the variational posterior over the latent space.
Next, we sample a latent variable $\mathbf{z}$ from the variational posterior using the reparametrization trick.
We then alter each of the $J$ dimensions of the sampled $\mathbf{z}$: the value of every latent dimension $z_j$ is first zeroed and then traversed in the range $[-5, 5]$ in ten steps while keeping all the values of all the others latent dimensions $z_{\neq j}$ fixed.
The updated $\mathbf{z}$ is then passed to the decoder to generate a new digit image that reflects the effect in image space of the perturbation we made to the data in latent space.

%% file: sections/04_results.tex
%%%%%%%%%%%%%%%%%%%%%%%%%%%%%%%%%%%%%%
%% Results
%%%%%%%%%%%%%%%%%%%%%%%%%%%%%%%%%%%%%%

\section{Results}
\label{results}

\subsection{Standard VAE}
\label{sec:results:vae}

\begin{figure}[!t]
  \centering
  \begin{subfigure}[b]{0.49\columnwidth}
    \centering
    \includegraphics[width=\textwidth]{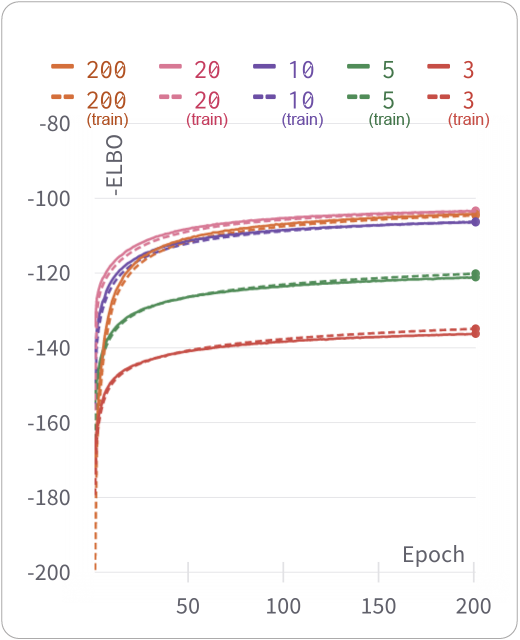}
    \caption{ELBO}
    \label{fig:vae:elbo}
  \end{subfigure}
  \begin{subfigure}[b]{0.49\columnwidth}
    \centering
    \includegraphics[width=\textwidth]{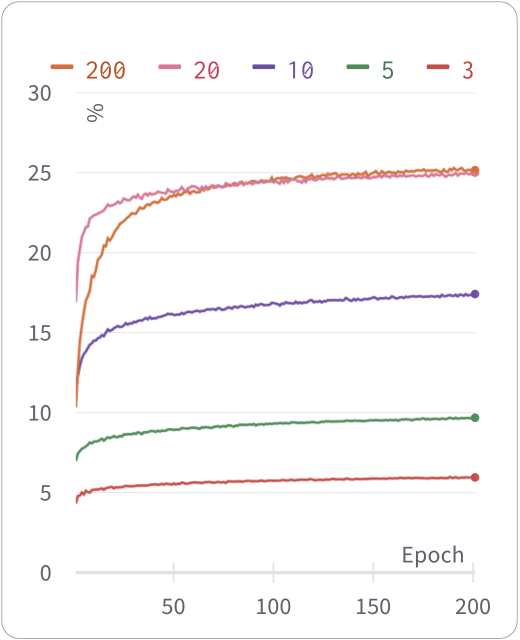}
    \caption{Ratio KL Divergence / ELBO}
    \label{fig:vae:elbo:kl}
  \end{subfigure}
\caption{\textbf{Standard VAE}. We reproduced of a portion of the experiments in \cite{kingma_autoencodingvariational_2014} to validate our machine learning pipeline.
We trained a standard VAE for 200 epochs and tested the effect of changing the number of dimensions of the latent space $\mathbf{z}$ from $J=3$ until $J=200$ in both train and test datasets.
The plots show that the ELBO saturated well within the defined number of epochs in all four cases without overfitting.}
\end{figure}

We reproduce a portion of the experiments on presented in \cite{kingma_autoencodingvariational_2014}.
We gradually increase the size of the latent variable $\mathbf{z}$, experimenting with dimension steps of $J \in \{3, 5, 10, 20, 200\}$, and observe what their impact was on the ELBO.

As Figure \ref{fig:vae:elbo} shows, there are only minimal improvements to the decreasing the value of the objective function after 10 latent dimensions, $J>10$.
The close gap between the training and testing curves in within the allocated VAE's training epochs suggest the trained VAEs does not overfit, even when the dimensionality of the latent space scales up to $J=200$.
This finding is in accordance with Kingma and Welling who attributed this phenomenon to the regularizing effect of the Gaussian prior and the KL divergence term in the ELBO \cite{kingma_autoencodingvariational_2014}.
However, latent variables with larger dimensionality raise the proportion that the KL divergence contributes to the ELBO.
This contribution oscillates between about 6\% when $J=3$ and 25\% when $J=200$ (see Figure \ref{fig:vae:elbo:kl}).
The increase in the contribution of the KL divergence term of the ELBO can be understood as a cumulative penalty accrued because the VAE is simultaneously fitting more dimensions of $\mathbf{z}$ to the Gaussian prior.

\subsection{$\beta$-VAE}

\begin{figure}[!t]
  \centering
  \begin{subfigure}[b]{0.49\columnwidth}
    \centering
    \includegraphics[width=\textwidth]{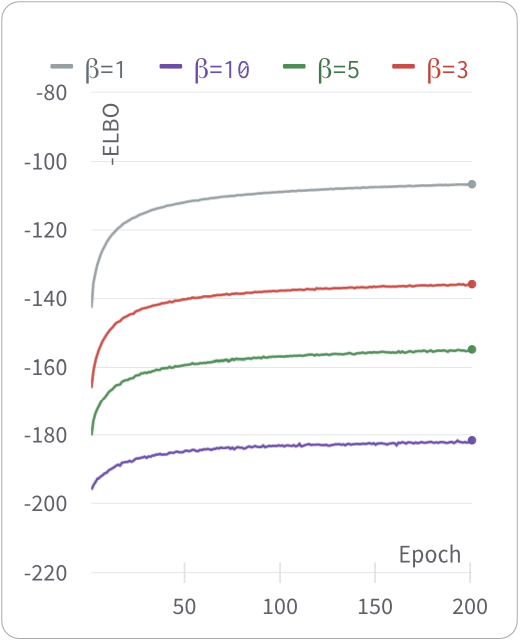}
    \caption{ELBO}
    \label{fig:betavae:elbo}
  \end{subfigure}
  \begin{subfigure}[b]{0.49\columnwidth}
    \centering
    \includegraphics[width=\textwidth]{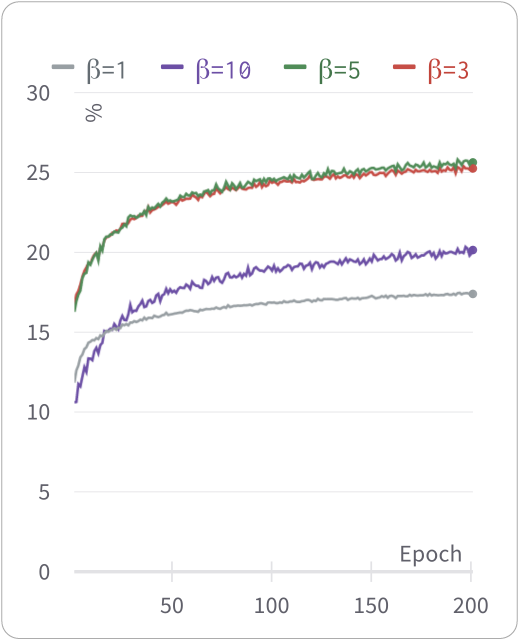}
    \caption{Ratio KL Divergence / ELBO}
    \label{fig:betavae:elbo:kl}
  \end{subfigure}
\caption{$\mathbf{\beta}$\textbf{-VAE}. Sensitivity study on the effect that different values of $\beta$ have on the ELBO.
In Figure \ref{fig:betavae:elbo:kl}, the participation of the KL divergence in the ELBO is lower when $\beta=10$ than when $\beta \in \{3, 5\}$.
}
\label{fig:betavae:plots}
\end{figure}

We carry out a sensitivity analysis on the hyperparameter $\beta$ by training a $\beta$-VAE with different values of $\beta \in \{1, 3, 5, 10, 20\}$.
We fix the number of latent dimensions to $J=10$ in all the experiments hereafter for the $\beta$-VAE as further increases to $J$ did not lead to decreasing the negative value of the ELBO as we report in Section \ref{sec:results:vae}).
Figure \ref{fig:betavae:elbo} shows higher values of $\beta$ lead to larger negative ELBO.
The value of the ELBO after 200 epochs decreases by almost 70\% when $\beta=10$ compared to when $\beta=1$.
This result suggests that further increasing $\beta$ would lead to a poorer log-likelihood estimate of the data.
In contrast, we observe that the contribution of the KL divergence between the variational posterior $q_{\phi}(\mathbf{z}|\mathbf{x})$ and the Gaussian prior $p(\mathbf{z})$ to the ELBO diminishes as $\beta$ is larger, except for when $\beta=10$ (see Figure \ref{fig:betavae:elbo:kl} ).

\begin{figure}[!b]
  \centering
  \begin{subfigure}[b]{0.6\columnwidth}
    \centering
    \includegraphics[width=\textwidth]{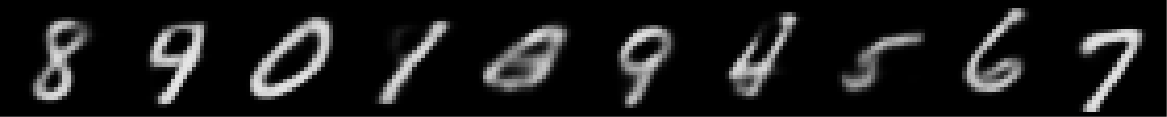}
    \caption*{Input data}
  \end{subfigure}
  \begin{subfigure}[b]{0.6\columnwidth}
    \centering
    \vspace{0.5mm}
    \includegraphics[width=\textwidth]{figures/mnist-strip-beta-1-01.png}
    \caption*{$\beta=1$}
  \end{subfigure}
  \begin{subfigure}[b]{0.6\columnwidth}
    \centering
    \vspace{0.5mm}
    \includegraphics[width=\textwidth]{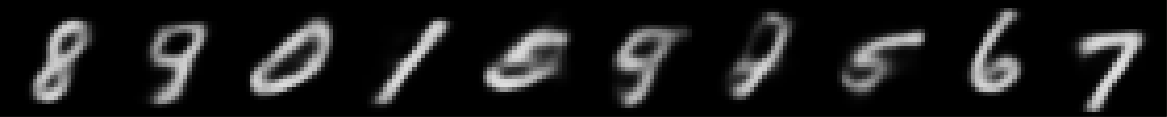}
    \caption*{$\beta=5$}
  \end{subfigure}
  \begin{subfigure}[b]{0.6\columnwidth}
    \centering
    \vspace{0.5mm}
    \includegraphics[width=\textwidth]{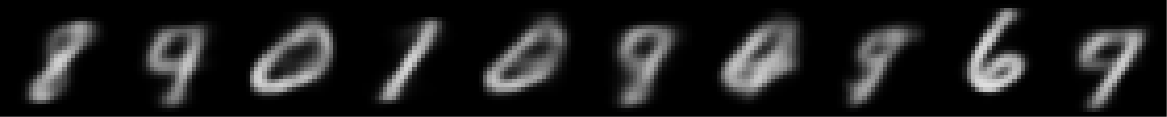}
    \caption*{$\beta=10$}
  \end{subfigure}
  \begin{subfigure}[b]{0.6\columnwidth}
    \centering
    \vspace{0.5mm}
    \includegraphics[width=\textwidth]{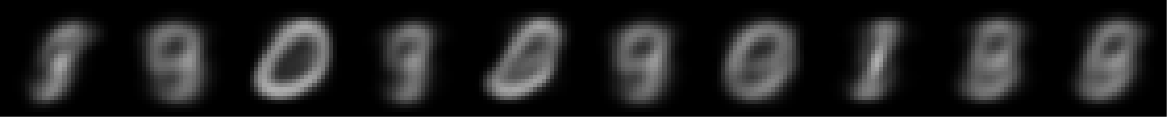}
    \caption*{$\beta=20$}
  \end{subfigure}
  \caption{\textbf{Digit reconstruction}. How does the choice of $\beta$ affect the visual aspect of the hand-written digits decoded by a $\beta$-VAE? Higher values of $\beta$ are conducive to blurrier reconstructions.}
  \label{fig:betavae:reconstruction}
\end{figure}

We also study the effect that increasing $\beta$ has on the visual quality of the digits reconstructed by the decoder of the $\beta$-VAE (see Figure \ref{fig:betavae:reconstruction}).
In contrast to the standard VAE (i.e. equivalent to setting $\beta=1$), the decoded images are consistently blurrier but still preserved the key features.
However, digits visually unidentifiable when $\beta=20$.
Figure \ref{fig:betavae:grid:8} shows that visual patterns begin to emerge when we traverse the latents with $\beta=10$.
For example, the latent dimensions $z_{2}$, $z_{3}$, $z_{6}$, $z_{10}$ do not experience any changes after traversing them, which suggests that the $\beta$-VAE model started to concentrate data variance over particular latent directions and neutralize information along others.
However, the visual recognition of single generative factors corresponding to single feature dimensions is still unclear.
For example, in Figure \ref{fig:betavae:grid:8}, traversing single latent dimensions results in inter-digit transformations instead of modifying any intrinsic visual property: digit 8 transitioned between digit 6 ($z_9=-5$) when and digit 3 ($z_9=5$) when traversing the latent space along dimension $z_9$.

%%%
\subsection{Conditional $\beta$-VAE}

\begin{figure}[!t]
  \centering
  \begin{subfigure}[b]{0.49\columnwidth}
    \centering
    \includegraphics[width=\textwidth]{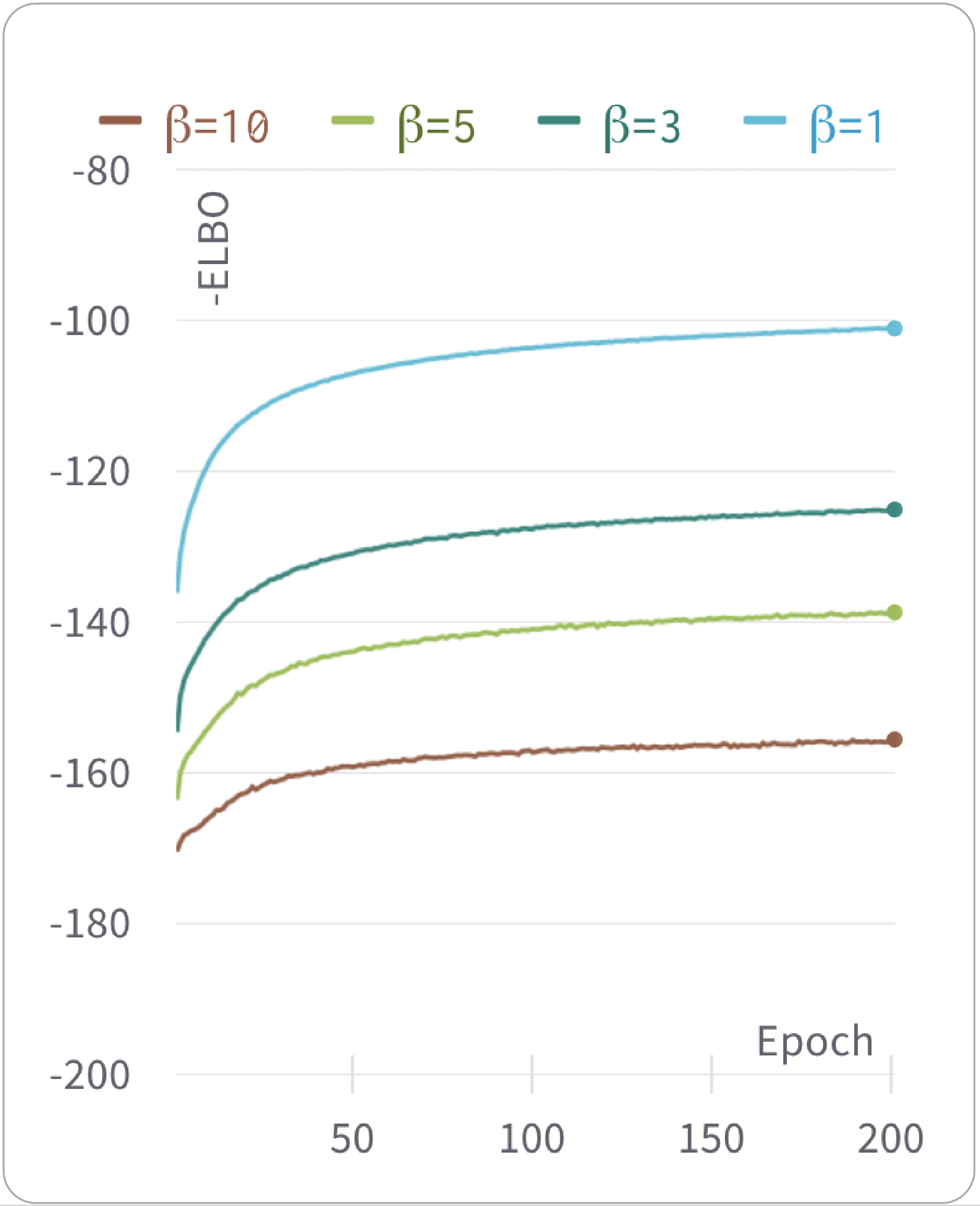}
    \caption{ELBO}
    \label{fig:cond:betavae:elbo}
  \end{subfigure}
  \begin{subfigure}[b]{0.49\columnwidth}
    \centering
    \includegraphics[width=\textwidth]{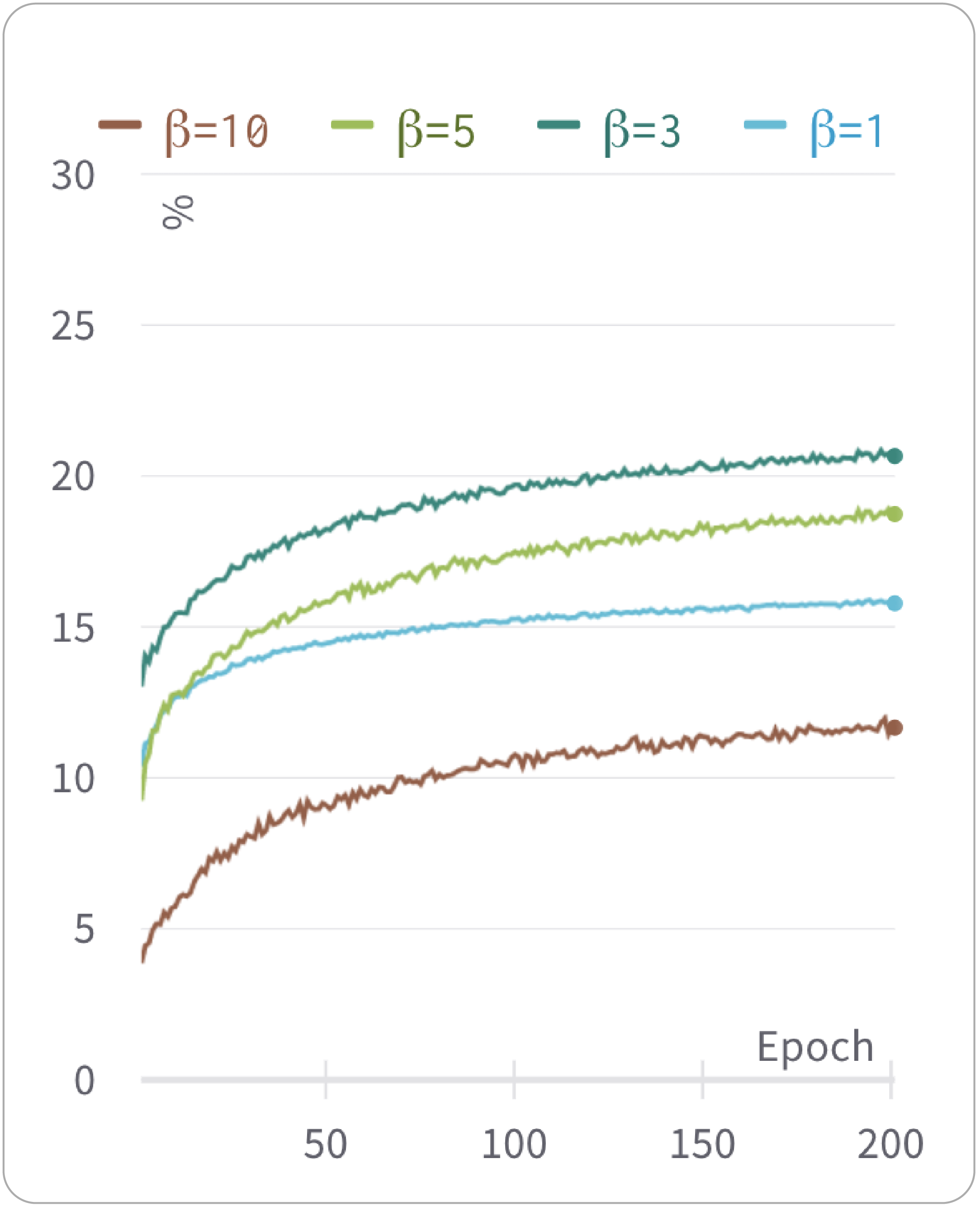}
    \caption{Ratio KL Divergence / ELBO}
    \label{fig:cond:betavae:elbo:kl}
  \end{subfigure}
\caption{\textbf{Conditional $\beta$-VAE}.
How does the addition of labels to the input dataset affect the learning process of a $\beta$-VAE? 
Overall, the minimization of the negative ELBO follows a similar downward trend than that of the unconditioned $\beta$-VAE for the four values of $\beta$ we test, $\beta \in \{1, 3, 5, 10\}$.
}
\end{figure}

We study the effect that using the class labels had on disentangling the latent space learned by a $\beta$-VAE.
We set the number of dimensions of the latent vector $\mathbf{z}$ in all the experiments we describe in this section to $J=10$, but we monotonically increase the value of $\beta$ in the range $\beta \in \{1, 3, 5, 10\}$.
The VAE ingests data pairs $(\mathbf{x}, \mathbf{u})$, where $\mathbf{u}$ is the digit class label. 
The class label $\mathbf{u}$ is encoded as a one-hot vector of size ten that is concatenated with the 784-dimensional original input vector.
We also concatenate the class one-hot vector to the sampled latent vector $\mathbf{z}$ that is input to the VAE decoder.
Therefore, we adjust the architecture of the VAE accordingly and add ten more units to the input layers of both the encoder and decoder: the number of of units in the first layer of the encoder increases from $784$ to $794$ whereas the number of units in the first layer of the decoder is $J+10=20$.

Figure \ref{fig:cond:betavae:elbo} shows that the ELBO and the KL divergence contribution fluctuates in agreement with the unconditional $\beta$-VAE, regardless of the choice of $\beta$: higher $\beta$ values lead to a lower ELBO estimate.
The KL Divergence over ELBO ratio was lowest when $\beta=10$, even lower than the case where $\beta=1$.

The figure also helps identifying that adding the class labels increase the fit of the model to the data.
The negative of the approximation to the log marginal likelihood is consistently 15\% lower for the conditioned $\beta$-VAE than it is in the unconditioned case.
For example, when $\beta=5$, the negative of the ELBO is -160 in the former case (Figure \ref{fig:betavae:elbo}), whereas this figure decreases to -140 in the latter (Figure \ref{fig:cond:betavae:elbo}).
Moreover, when $\beta \in \{1, 3, 5\}$, Figure \ref{fig:cond:betavae:elbo:kl} tells us that the participation of the KL divergence in the ELBO increases steadily from 15\% until over to 20\%, but it interestingly sees a sharp decline when $\beta=10$.

After training the $\beta$-VAE on the labeled dataset, we observe that the disentanglement of the latent space is qualitatively clearer compared to that produced by the $\beta$-VAE framework trained on the unlabeled dataset with $\beta=10$.
As we show in Figure \ref{fig:cond:betavae:grid:8}, one of the most significant findings is that after running inference on images of the digits, seven of the ten latent dimensions $z_i$ are not changed by the latent traversals.
The second most significant result is that dimensions $z_6$, $z_9$, and $z_{10}$ concentrate all the information related to the digit's generative factors, which we associate to the latent directions that control the line-weight, lateral tilt, and width of the hand-written digits.
These disentangled latent directions remain consistent even as we traverse the latent space of the conditional $\beta$-VAE for all nine digits (see Figure \ref{fig:disentangled:digits}).

\subsection{Comparison: $\beta$-VAE and conditional $\beta$-VAE}

\begin{figure}[!t]
  \centering
  \begin{subfigure}[b]{0.49\columnwidth}
    \centering
    \includegraphics[width=\textwidth]{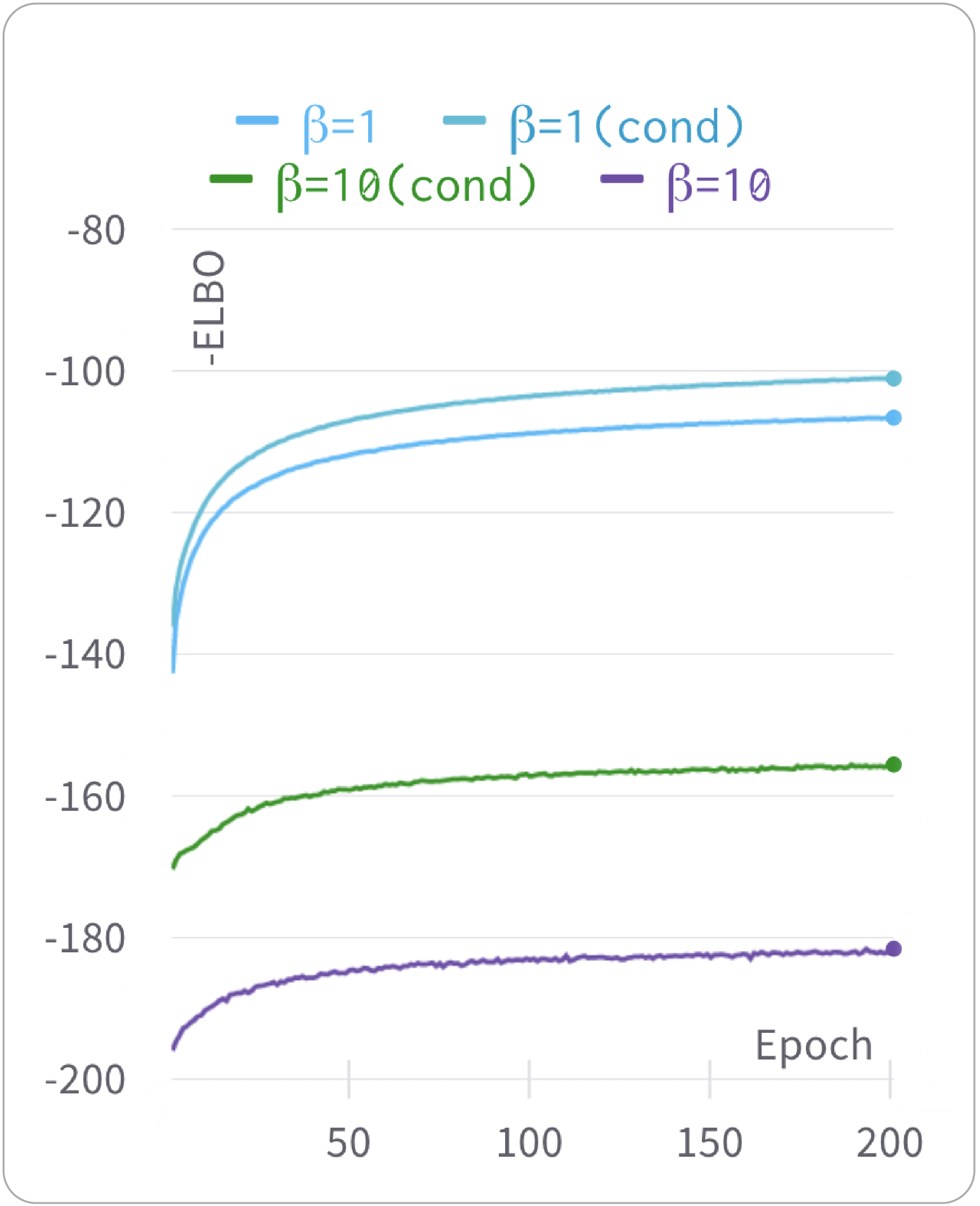}
    \caption{ELBO}
    \label{fig:comp:betavae:elbo}
  \end{subfigure}
  \begin{subfigure}[b]{0.49\columnwidth}
    \centering
    \includegraphics[width=\textwidth]{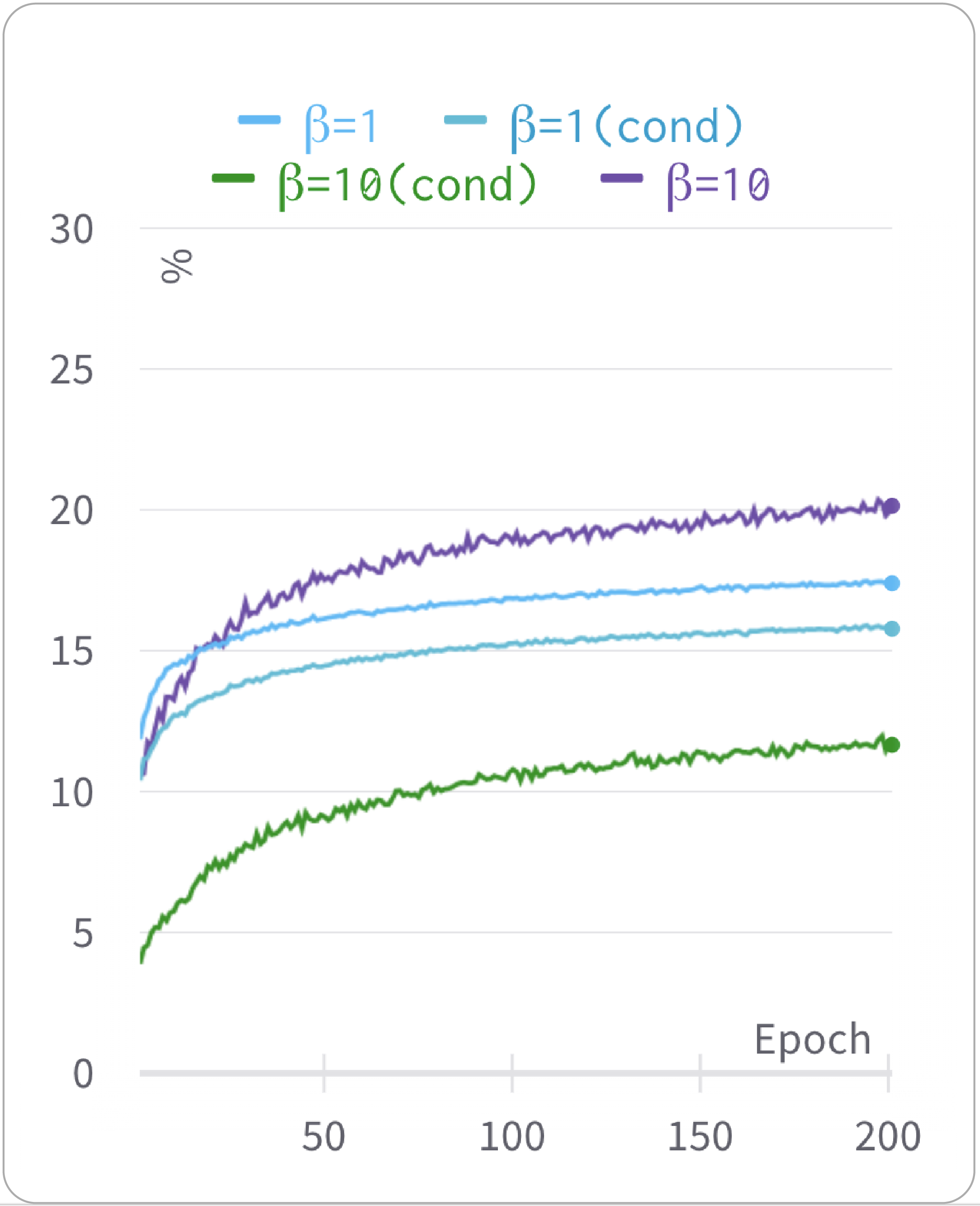}
    \caption{Ratio KL Divergence / ELBO}
    \label{fig:comp:betavae:elbo:kl}
  \end{subfigure}
\caption{\textbf{Standard VAE vs. $\beta$-VAE vs. conditional $\beta$-VAE}.
Metrics comparison between a standard VAE ($\beta=1$) \cite{kingma_autoencodingvariational_2014} and a $\beta$-VAE ($\beta=10$) \cite{higgins_betavaelearning_2016} with unlabeled and labeled data. $J=10$ in the all four tests.
}
\end{figure}

We examine and compare the behavior of $\beta$-VAE, in both the conditioned and unconditioned case with two different reference values $\beta=1$ and $\beta=10$.
In terms of the magnitude of the ELBO, the conditioned dataset with $\beta=1$ exhibits the highest value and the unconditioned case with $\beta=10$ the lowest, as depicted in Figure \ref{fig:comp:betavae:elbo}.
This concur with our previous experimental results that show that higher values of $\beta$ lead to lower values of the ELBO.

One of the most interesting findings pertains to the KL divergence to ELBO ratio, where conditioning the dataset leads to changes in the trend we observed in previous experiments.
In Figure \ref{fig:comp:betavae:elbo:kl}, for example, the contribution of the KL divergence to the ELBO is almost cut from 20\% down to almost 10\% when $\beta=10$, 
Since the class-labeled dataset with this value of $\beta$ produces the best disentanglement results so far, we hypothesize whether minimizing the participation of the KL divergence in the ELBO calculation while preserving a reasonable reconstruction error is conducive to good latent space disentanglement.
A more detailed investigation on the matter is left to future work.

%% file: sections/05_conclusion.tex
%%%%%%%%%%%%%%%%%%%%%%%%%%%%%%%%%%%%%%
%% Conclusion
%%%%%%%%%%%%%%%%%%%%%%%%%%%%%%%%%%%%%%

\section{Conclusion}
\label{conclusion}

\begin{figure*}[!t]
  \centering
  \begin{subfigure}[b]{0.32\textwidth}
    \centering
    \includegraphics[width=\textwidth]{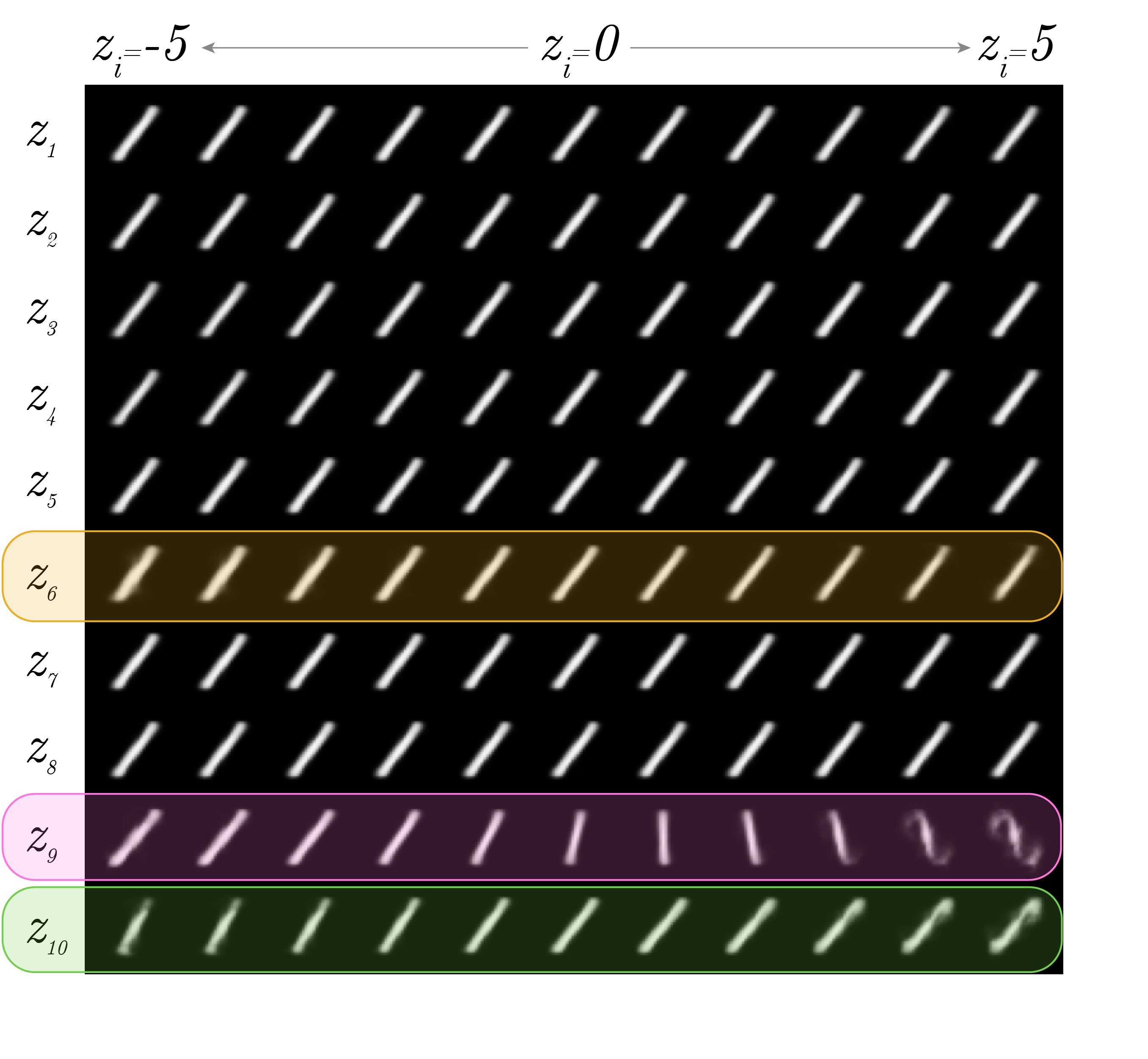}
  \end{subfigure}
  \begin{subfigure}[b]{0.32\textwidth}
    \centering
    \includegraphics[width=\textwidth]{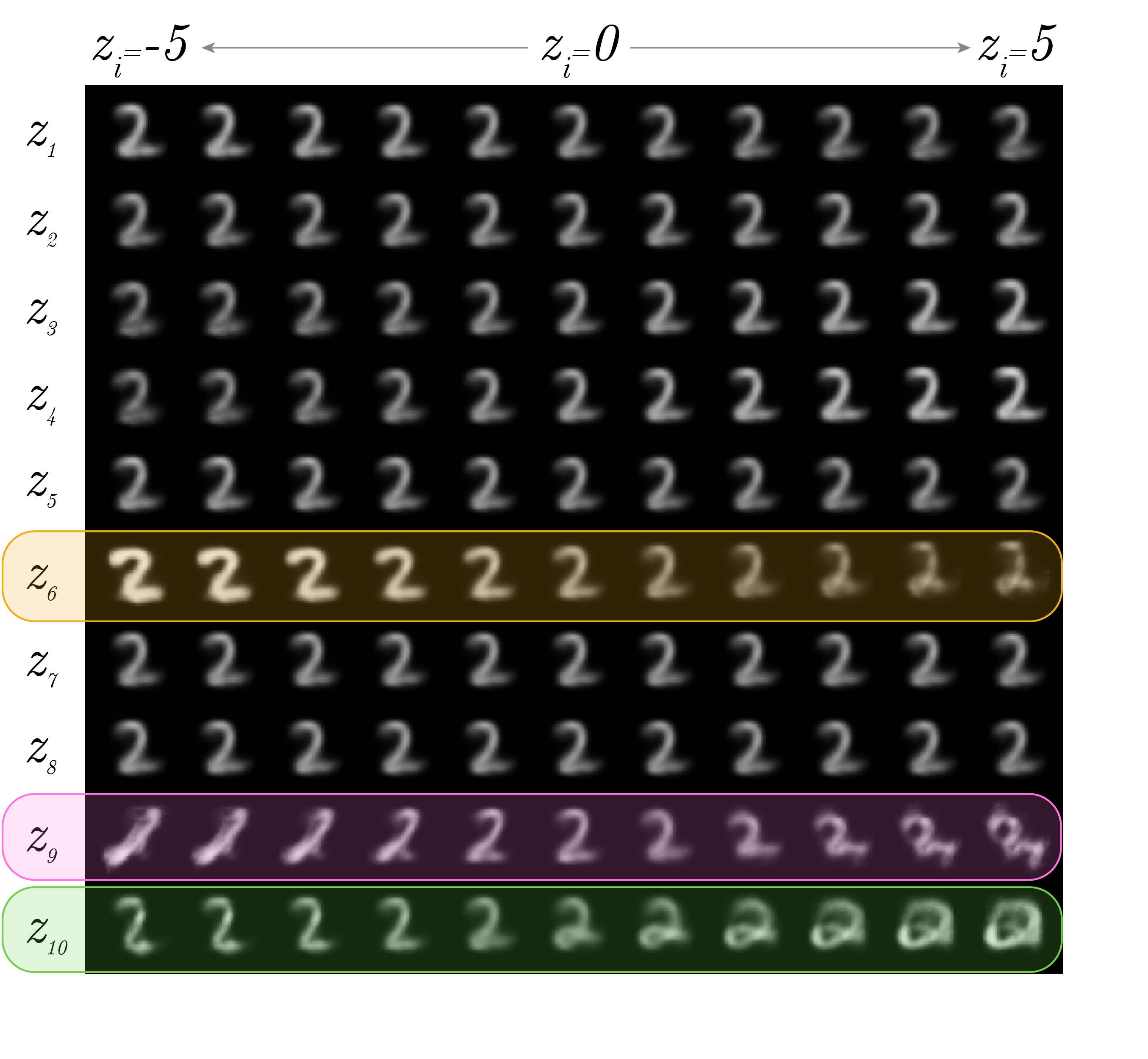}
  \end{subfigure}
  \begin{subfigure}[b]{0.32\textwidth}
    \centering
    \includegraphics[width=\textwidth]{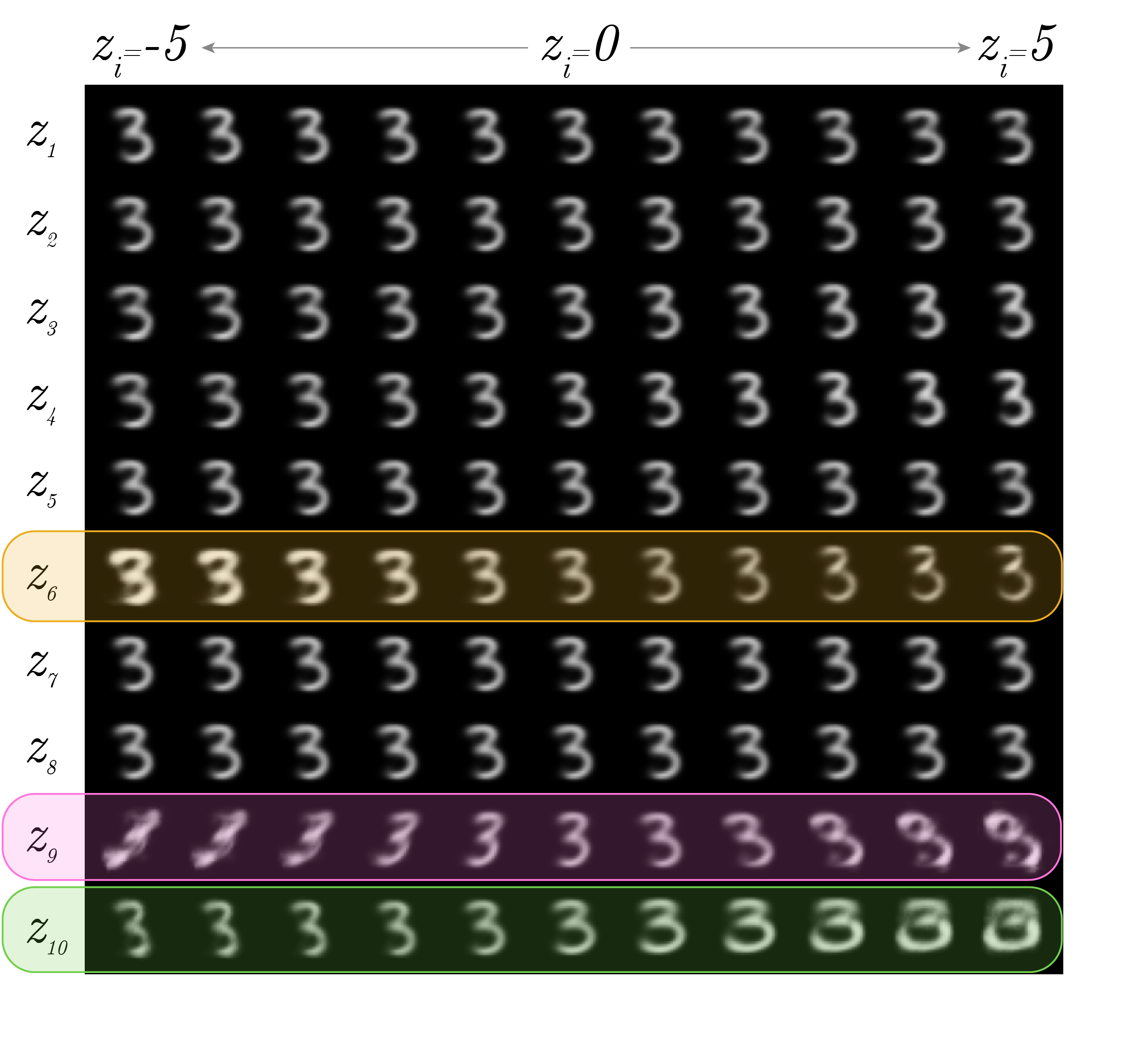}
  \end{subfigure}
  \begin{subfigure}[b]{0.32\textwidth}
    \centering
    \includegraphics[width=\textwidth]{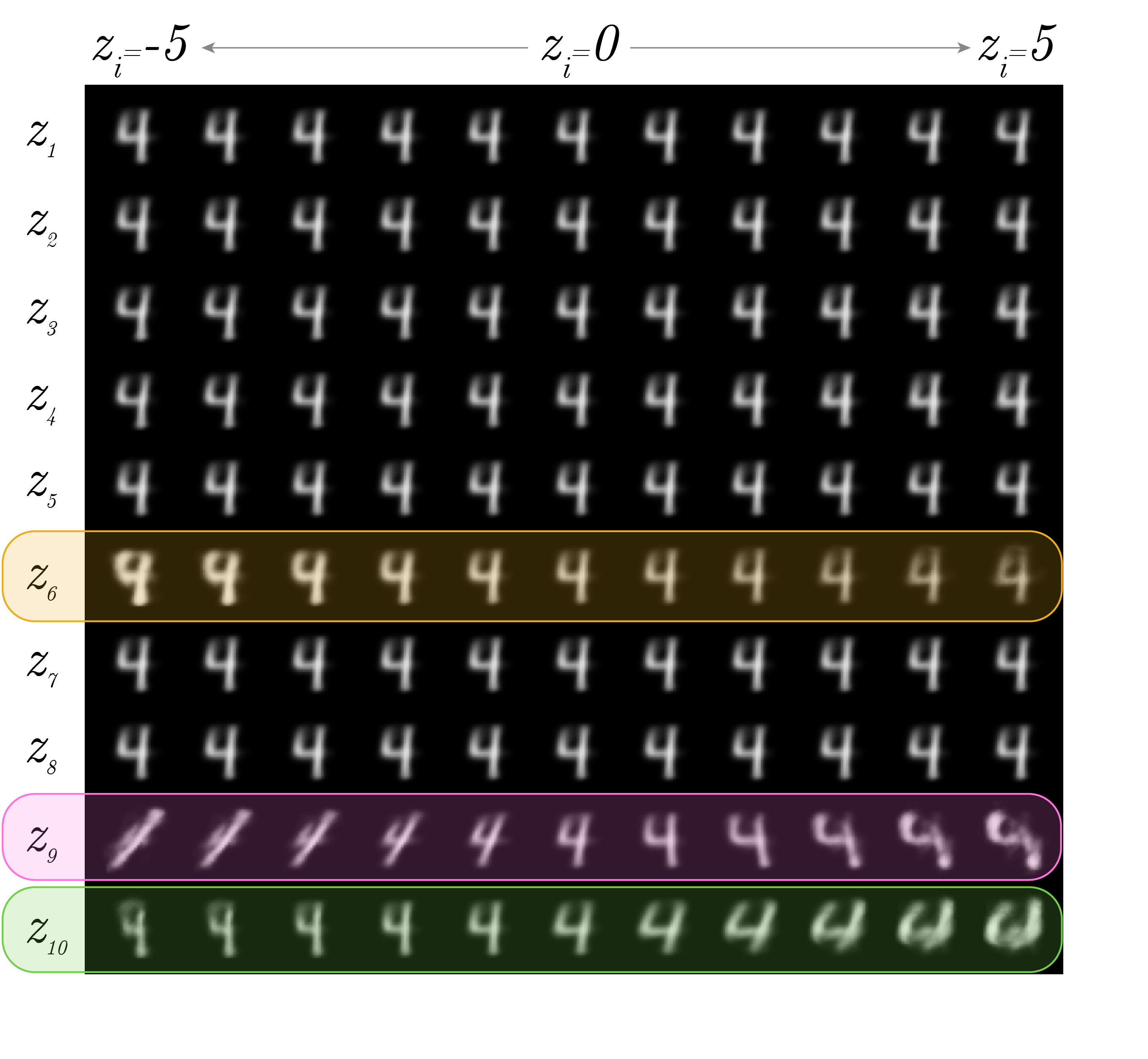}
  \end{subfigure}
  \begin{subfigure}[b]{0.32\textwidth}
    \centering
    \includegraphics[width=\textwidth]{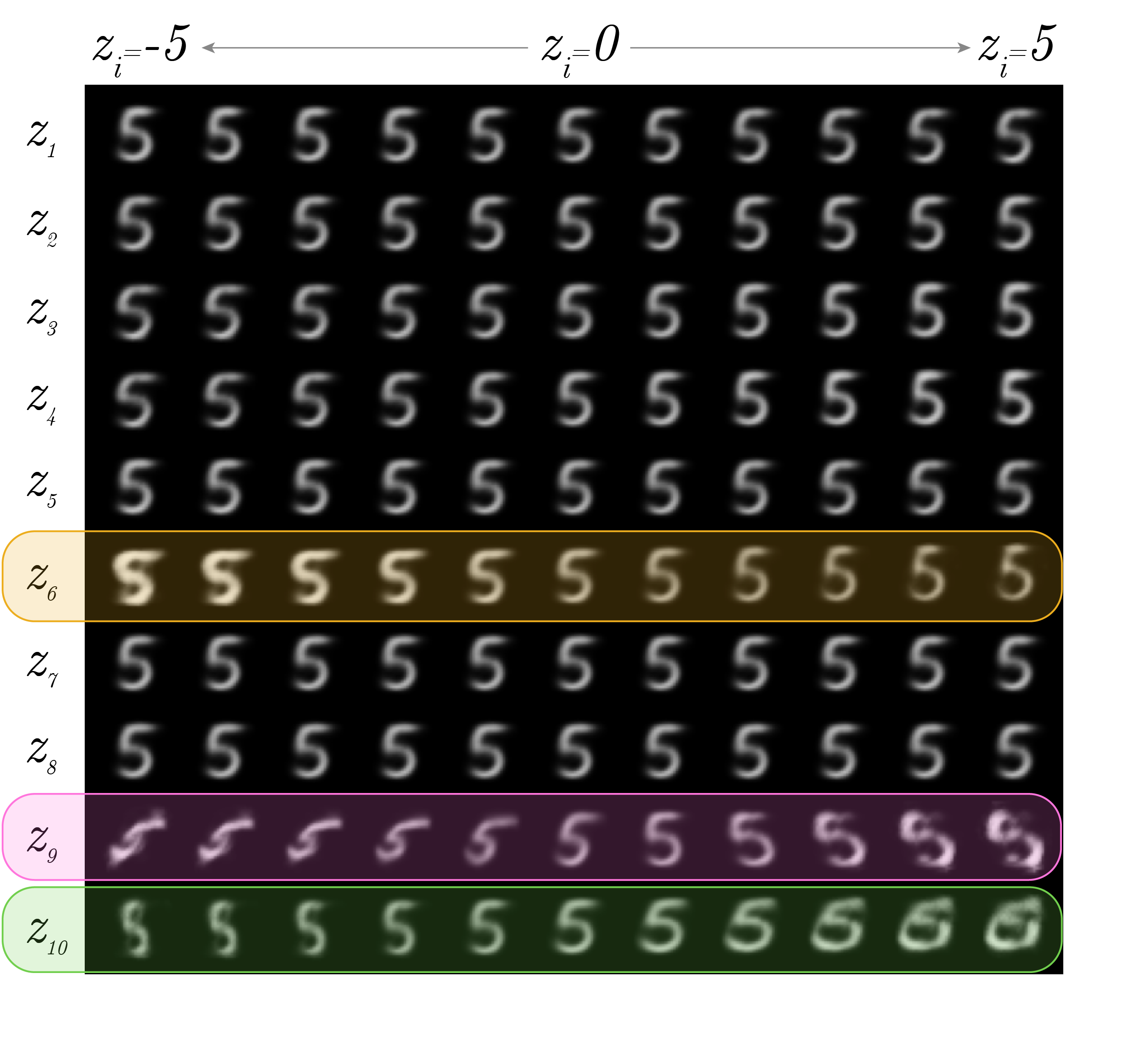}
  \end{subfigure}
  \begin{subfigure}[b]{0.32\textwidth}
    \centering
    \includegraphics[width=\textwidth]{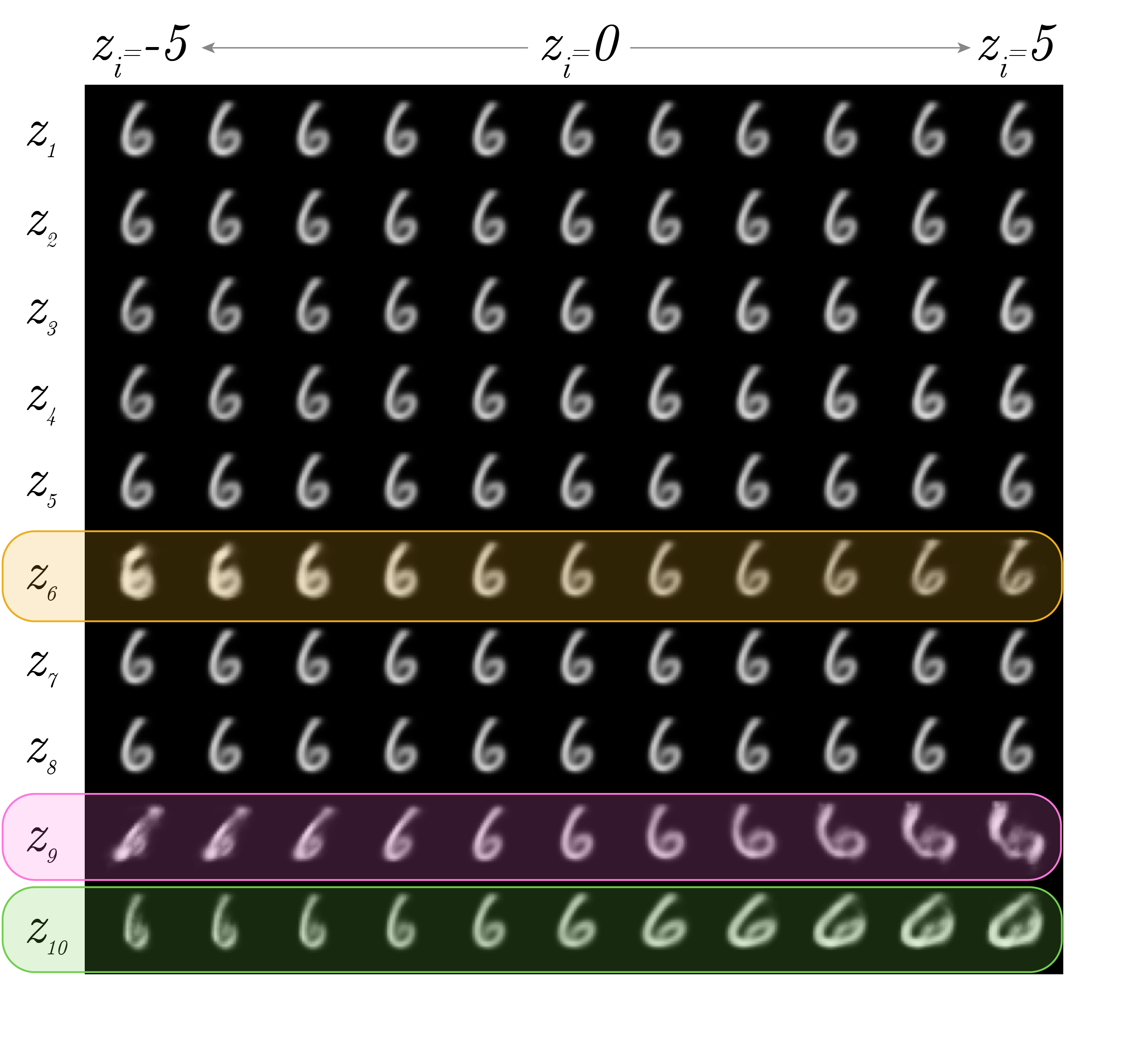}
  \end{subfigure}
  \begin{subfigure}[b]{0.32\textwidth}
    \centering
    \includegraphics[width=\textwidth]{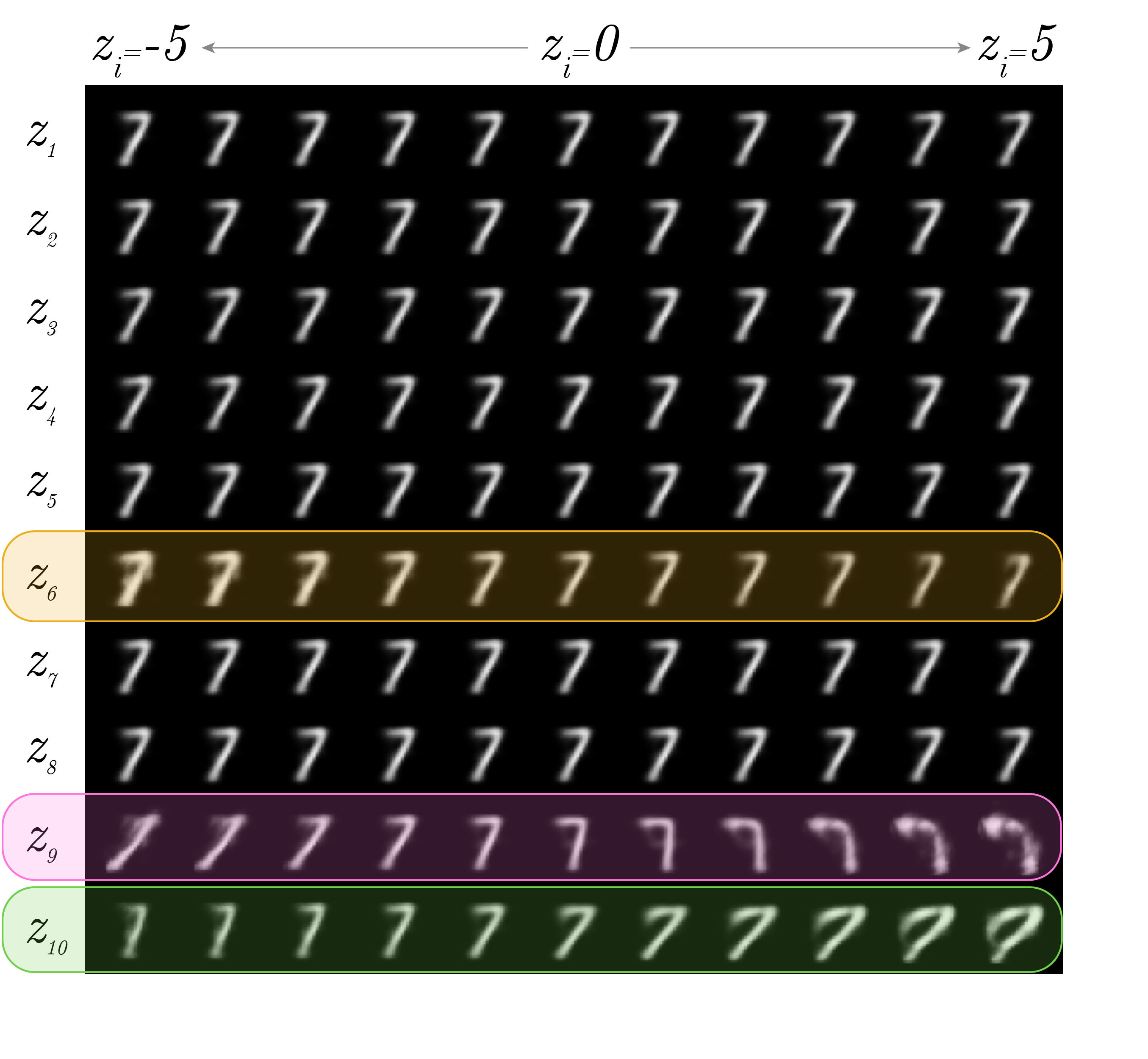}
  \end{subfigure}
  \begin{subfigure}[b]{0.32\textwidth}
    \centering
    \includegraphics[width=\textwidth]{figures/cond_beta_8_5-01.png}
  \end{subfigure}
  \begin{subfigure}[b]{0.32\textwidth}
    \centering
    \includegraphics[width=\textwidth]{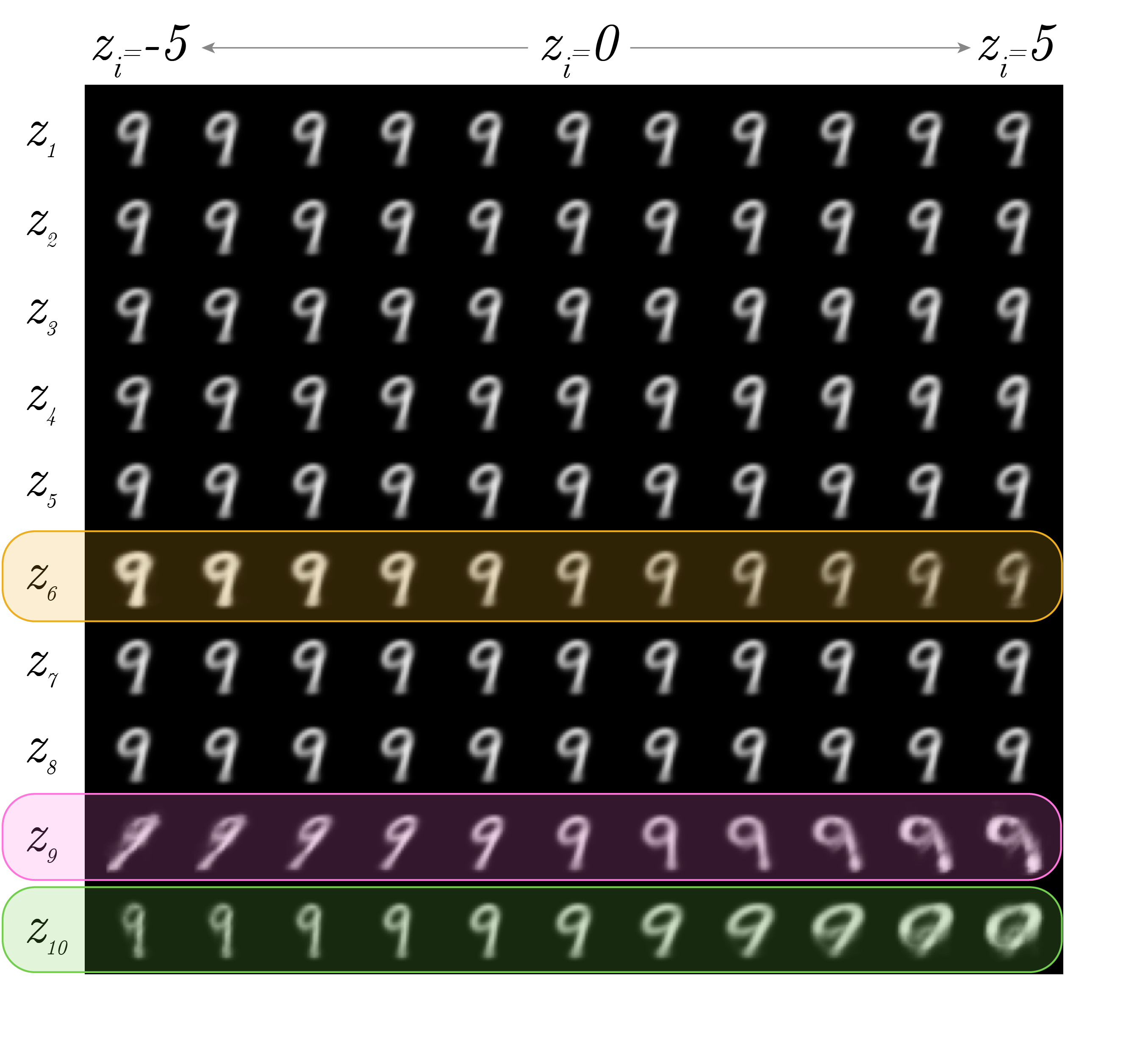}
  \end{subfigure}
\caption{\textbf{Disentangled latent space}.
The alignment between line weight, tilt and digit width and the latent dimensions $z_6$, $z_9$ and $z_{10}$ we found after traversing the latent space learned by the conditional $\beta$-VAE remained consistent across all nine digits.}
\label{fig:disentangled:digits}
\end{figure*}

In this paper, we implemented and trained three different VAEs to study latent space disentanglement on a dataset of 60,000 images of hand-written digits: a standard VAE \cite{kingma_autoencodingvariational_2014}, a $\beta$-VAE \cite{higgins_betavaelearning_2016} and a conditional $\beta$-VAE \cite{locatello_challengingcommon_2019}.
We analyzed the effect of varying the magnitude of the hyperparameter $\beta$, and that of conditioning images on discrete class labels $\mathbf{u}$ on the disentanglement of the latents.

We found that using $\beta=10$ together with label-conditioned data led to the clearest level of disentanglement through our experiments, revealing single latent directions that allowed for individual control of three generative factors in the images supplied to the VAE.
These generative factors corresponded to the line weight, the tilt and the width of the digit images.
This finding supports the idea that good disentanglement is contingent upon some a priori level of supervision on the input data \cite{locatello_challengingcommon_2019}.
However, our experiments also suggest that using a moderate value of $\beta$ to scale the KL divergence in the ELBO is also instrumental to arrive at a disentangled latent space.

Disentangled representation learning is an active area of research with plenty of interesting challenges ahead.
Several routes to extend our work are thus outlined.
First, we plan to develop a consistent and robust quantitative disentanglement metric --resorting to qualitative visual heuristics to evaluate latent space disentanglement may be cumbersome.
We are interested in following a methodology similar to the prediction-based measurements exposed in \cite{higgins_betavaelearning_2016} to this end.
An alternative approach to disentanglement we would like to explore later is to directly perform \textit{independent component analysis} (ICA) on the learned latents $\mathbf{z}$ and to evaluate whether ICA facilitates finding disentangled directions without having to use the hyperparameter $\beta$ or any class labels, as suggested in recent work \cite{khemakhem_variationalautoencoders_2020}.

We also aim to leverage other neural networks in the encoder and the decoder components of the VAE to learn better-quality latents by exploiting the symmetries in the data, in particular, the translational invariance of image-structured data that we used herein.
One concrete example is to use convolutional neural networks instead of multilayer perceptrons.
Finally, we anticipate working with richer and structured prior distributions for the variational distribution, other than Gaussians, which may ultimately better capture the hidden and disentangled generative structure of the input data.